\documentclass[journal,onecolumn]{IEEEtran}
\usepackage{graphicx}
\usepackage{hyperref}

\usepackage{cite}

\ifCLASSINFOpdf
\else
\fi
\usepackage{amsmath}
\usepackage{algorithmic}

\usepackage{array}
\usepackage{url}

\begin{document}
%
\title{Bflier’s: A Novel Butterfly Inspired Multi-robotic Model in Search of Signal Sources}
%
%
%

\author{Chakravarthi~J$^{1}$, 
    Vinod~Babu~P$^{1}$, Pavan~B$^{2}$, Ashok~U$^{2}$,  \\ Marek~Kolencik$^{3}$, Martin~Šebesta$^{4}$, Ramakanth~Illa$^{5}$ \\
$^{1}$Rajiv~Gandhi~University~of~Knowledge~Technologies - Nuzvid \\ 
$^{2}$International~Institute~of~Information~Technology - Hyderabad \\ 
$^{3}$Slovak~University~of~Agriculture \\ 
$^{4}$Comenius~University~Bratislava \\
 $^{5}$Vellore~Institute~of~Technology 
}
\maketitle

\begin{abstract}
The diversified ecology in nature had various forms of swarm behaviors in many species. The butterfly species is one of the prominent and a bit insightful in their random flights and converting that into an artificial metaphor would lead to enormous possibilities. This paper considers one such metaphor known as Butterfly Mating Optimization (BMO). In BMO, the Bfly follows the patrolling mating phenomena and simultaneously captures all the local optima of multimodal functions. To imitate this algorithm, a mobile robot (Bflybot) was designed to meet the features of the Bfly in the BMO algorithm. Also, the multi-Bflybot swarm is designed to act like butterflies in nature and follow the algorithm's rules. The real-time experiments were performed on the BMO algorithm in the multi-robotic arena and considered the signal source as the light source. Subsequently, various strategies are annotated while Bflybots are converged towards multiple signal sources. The experimental results show that the BMO algorithm is applicable to detect multiple signal sources with significant variations in their movements i.e., static and dynamic. In the case of static signal sources, with varying initial locations of Bflybots, the convergence is affected in terms of time and smoothness. Whereas the experiments with varying step-size leads to their variation in the execution time and speed of the bots.  
In this work, experiments were performed in a dynamic environment where the movement of the signal source in both maneuvering and non-maneuvering scenarios. The Bflybot swarm is able to detect the single and multi-signal sources, moving linearly in between two fixed points, in circular, up and down movements. In the case of circular movement, the bots exhibit the to and fro motion after completing a certain number of iterations. To evaluate the BMO phenomenon, various ongoing and prospective works such as mid-sea ship detection, aerial search applications, and earthquake prediction were discussed.
\end{abstract}


%

\section{Introduction}
%
%
%
%

 
Nature is the resource for a variety of species. All the species are doing complex activities with simple and elegant rules for their survival. The collective behavior of individual species is always a source of inspiration for scientific society. For example, though tiny in size ants apply a cooperative mechanism for foraging and nest building. On the same line of thought, to avoid predators fish schooling will apply extraordinary maneuvering. There are other species like birds to travel longer distances and honeybees for collection nectar, which applies collection behavior.  One branch of that collective cooperation is swarm intelligence \cite{b11} all agents pose the same rules and capabilities. This equality along with local cooperation in the agents tremendously leads to achieving global results.

A decade ago, one school of researchers have studied these behaviors and proposed artificial paradigms, such as PSO - Particle Swarm Optimization \cite{b1}, ACO - Ant Colony Optimization \cite{b2}, GSO - Glowworm Swarm Optimization \cite{b3} etc, are prominent among them. Among those, some of the algorithms were effective in detecting global optima and others are capable of locating local optima but the core aim of all these algorithms is to solve the problem with collective behavior which is called swarm behavior. Another school of researchers used these models and designed robotic platforms to detect (locate) multiple signal sources such as light, fire, plume, odour etc. The Kinbot \cite{b3} is one such recent experiment. In the same line of thought, this work presents the proposed butterfly metaphor \cite{b4}, simulations and multi-robotic experiments. This paper also presents several key ideas to implement the BMO metaphor on various real-time applications. 

\begin{figure}
    \centering
    \includegraphics[width=7in]{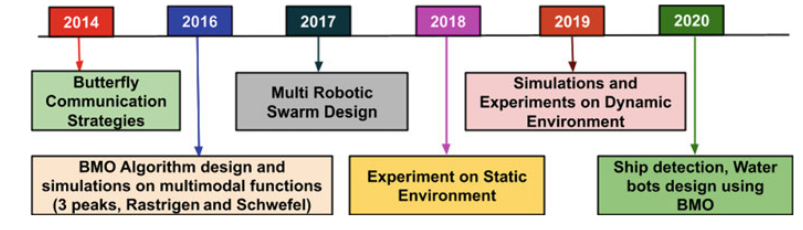}
    \caption{Work flow of butterfly mating optimization}
    \label{fig:work_folow}
\end{figure}

Figure~\ref{fig:work_folow}, briefly describes the entire pipeline of the Butter mating optimization algorithm and its evaluation. Initial attempts were made towards analyzing the communication strategies of butterflies in nature. Later, the BMO algorithm was designed and various virtual simulations were performed to verify the efficacy of the proposed algorithm. While proposing the algorithm, various comparisons were made with the popularly known optimization algorithms (like PSO, GSO, ACO). To mimic the BMO algorithm in the robotic platform, Bflybot was designed and verified the efficacy of the design by detecting a single light source. In recent times, work progresses on designing multiple Bflybots to do source localization with multiple robots, all of them are similar in structural design. Progressively, experiments were performed to detect multiple same and different color sources. To detect the dynamically varying sources, a few experiments were performed while the light sources were moving via. straight line, circular and up-down movement. Current works concentrate on the application of the BMO algorithm in real-time scenarios.

This chapter discusses the importance of the proposed Butterfly Mating Algorithm for the detection/co-location of the multimodal functions. Section-\ref{sec:sec2} briefly describes the inspiration for the formulation of the Butterfly mating optimization algorithm to mimic the mating behavior of the butterflies in nature. In section-\ref{sec:sec3}, details the simulation results and the evaluation of the BMO algorithm to colocate the peaks in benchmark models (Three peaks, Schwefel and Rastrigin functions). The section-\ref{sec:sec4} presents the perspective, and ongoing works to check the efficacy of the BMO algorithm in real-time applications. Finally section-\ref{sec:sec5}, discuss the conclusion made by the observations.



\section{Butterflies as New Inspiration}
\label{sec:sec2}
\subsection{Butterfly Communication Strategies}
The essential mode of communication in butterflies mainly happens through either mating or defense. Two major forms of communications strategies are patrolling and perching. In case of patrolling, male butterflies continuously search for the female butterflies by absorbing the UV, based on the UV reflected by the female butterflies the corresponding male butterflies recognize the female butterflies. Color and odor are the main auxiliary traits in patrolling. In perching, male butterflies are completely stagnant by sitting at the hilltop. The female butterflies which enter into the region will mate with them. The main attracting parameters in perching are size and movement. Sometimes, butterflies use the defense mechanism to protect from predators. 

\begin{figure}
    \centering
    \includegraphics[width=7in]{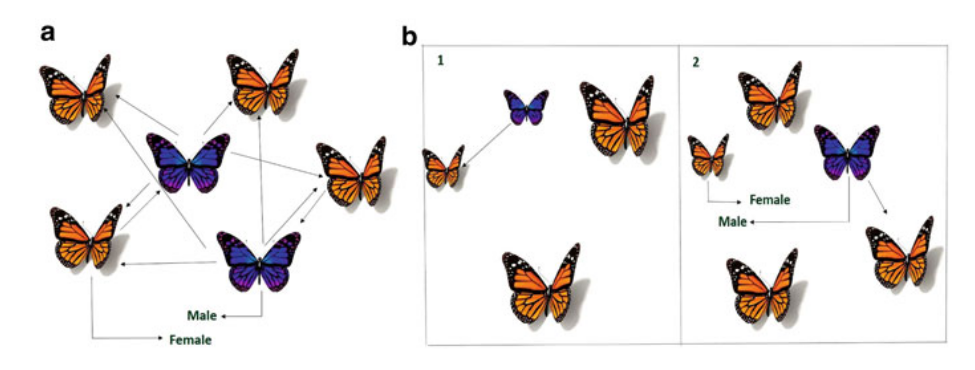}
    \caption{Pictorial representation (a) Patrolling (b) Perching}
    \label{fig:fig2}
\end{figure}

The bizarre flying of butterflies in nature leads to many queries. The state of art indicates that butterflies mainly do the communication for the sake of mating \cite{b5} \cite{b10}. Butterflies mate based on either patrolling (Fig~\ref{fig:fig2}.a) - (where the female butterflies reflect UV to male butterflies depends on distances, and the male will respond), or perching (Fig~\ref{fig:fig2}.b) - (where the male sits at the hill-top and search in its territory among females passing by, and further they use other traits such as size and fluttering) to select a mate. Based on these behaviours and virtual simulations done by sowmya et. al \cite{b5}, chakravarthi et. al \cite{b4} proposed Butterfly Mating Optimization (BMO) for simultaneously capturing all local optima of multimodal functions. This algorithm basically uses patrolling mating behaviour \cite{b10} and does not distinguish between male and female butterflies. Hence the ”butterfly in nature named as Bfly in the search space”. Various simulations were performed \cite{b5} to discard the discrimination between the male and female butterflies in the BMO algorithm. 

The four phases of the BMO algorithm are:

\subsection{UV Updation Phase}
\hspace{0.3cm} In this phase, \textit{UV} (Ultraviolet light) of each Bfly is updated in accordance with its fitness function value at the present location and is given by the following equation.
\begin{equation}
UV_{i}(t) = max\{0, b_{1}*UV_{i}(t-1) ~+ ~b_{2}*f(t)\}
\end{equation}
$UV$ is updated at every time index `$t$', giving more importance to the current fitness value \textit{f(t)} than the previous $UV$ value,  accordingly choose the constants $b_{1}$, $b_{2}$ such that $0 \leq b_{1} \le 1$ and $b_{2} > 1$.

\subsection{UV Distribution Phase}
\hspace{0.3cm}In this phase, each Bfly distributes its updated $UV$ value to remaining Bflies. This distribution mainly depends on distance between them such that the nearest Bfly receives  more $UV$ than the farthest one and helpful for Bflies to choose their \textit{l-mate} which is nearer to it and satisfies the eq. 3 \& 4 in \textit{l-mate} selection phase. For that an $i^{th}$ Bfly having intensity $UV_{i}$ reflects it's $UV$ value to the $j^{th}$ Bfly at a distance $d_{ij}$ which is given by

\begin{equation}
UV_{i \rightarrow j} = UV_{i} \times \frac{\displaystyle d_{ij}^{-1}}{\displaystyle \sum_{k}^{} d_{ik}^{-1}}
\end{equation}

where $i = 1,2,\ldots,N$; $N$ is the number of Bflies; $j = 1,2, \ldots, N$ and $j\neq i$; $UV_{i \rightarrow j }$ is $UV$ absorbed by $j^{th}$ Bfly from $i^{th}$ Bfly; d$_{ij}$ is the euclidean distance between $i^{th}$ and $j^{th}$ Bfly. $k= 1,2, \ldots,j, \ldots,N$ and $k\neq i$; d$_{ik}$ is the euclidean distance between $i^{th}$ and $k^{th}$ Bfly.

\subsection{L-mate Selection Phase}
\hspace{0.3cm}After \textit{UV} distribution, every Bfly is in search of its \textit{local mate}. Each Bfly has its own \textit{UV} and fitness values. For \textit{l-mate} selection, every $i^{th}$ Bfly arranges  the remaining Bflies in descending order based on their \textit{UV} values. The fitness value of $i^{th}$ Bfly is compared with remaining Bflies in that descending order. Whenever it finds more fitness value than its own for the first time, then it chooses the corresponding Bfly as its \textit{l-mate} and moves towards it. If any Bfly do not satisfies the below conditions(eq. 3 \& 4) then it has no \textit{l-mate} (means it is \textit{l-mate} to itself)\cite{b6}

\begin{equation}
 UV(i^{th}~Bfly) < UV(j^{th}~Bfly)
\end{equation}

\begin{equation}
 f(i^{th}~Bfly) < f(j^{th}~Bfly)
\end{equation}

where $i = 1,2,\ldots,N$; $j = 1,2,\ldots,N-1$, $N$ is the number of Bflies; $j$ is the index of Bflies in the descending order of $i^{th}$ Bfly.
\subsection{Movement Phase}
\hspace{0.3cm}After mate selection, each Bfly will move in the direction of its $l\mbox{-}mate$ with specified stepsize as follows.

\begin{equation}
 x_{i}(t+1) = x_{i}(t) ~+ ~B_{s}*\left\lbrace\frac{x_{l\mbox{-}mate}(t) - x_{i}(t)}{\| x_{l\mbox{-}mate}(t) - x_{i}(t)\|}\right\rbrace
\end{equation}

where $B_{s}$ is Bfly stepsize; $x_{i}(t)$ is the position of $i^{th}$ Bfly in a time index $t$.

\section{Simulations and Robotic Experiments}
\label{sec:sec3}
\subsection{ Bflies to capture simultaneous peaks in multimodal functions}

\begin{figure}
    \centering
    \includegraphics[width=7in]{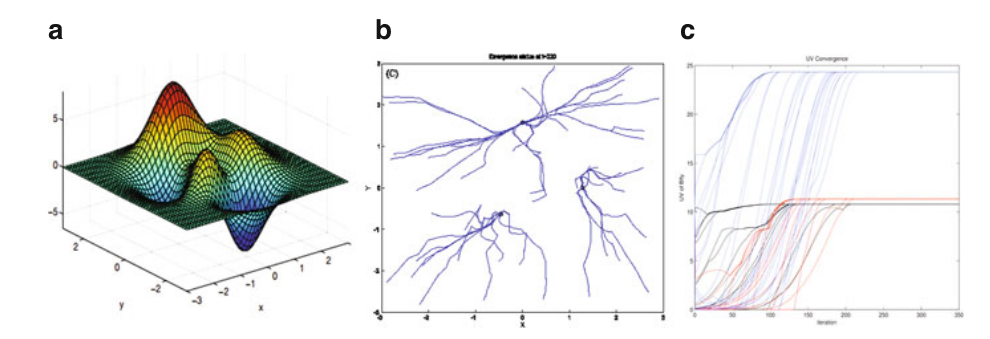}
    \caption{ (a) Three peaks function 	 (b) Emergence of Bflies 	         (c) UV convergence}
    \label{fig:fig3}
\end{figure}

\begin{figure}
    \centering
    \includegraphics[width=7in]{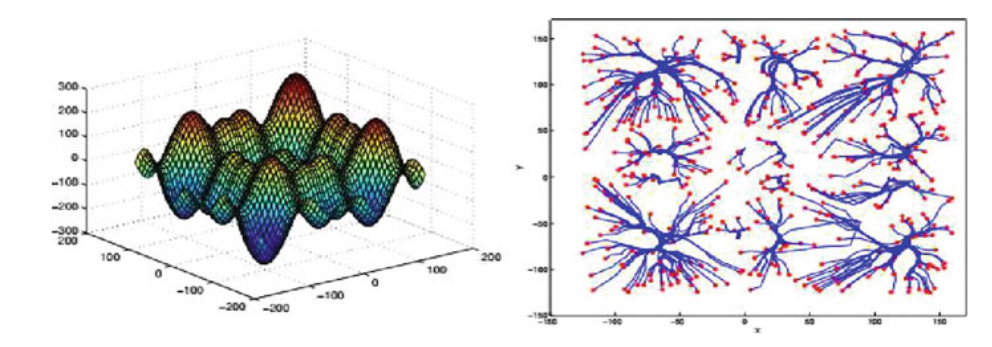}
    \caption{Schwefel function Convergence and Emergence plot}
    \label{fig:fig4}
\end{figure}

\begin{figure}
    \centering
    \includegraphics[width=7in]{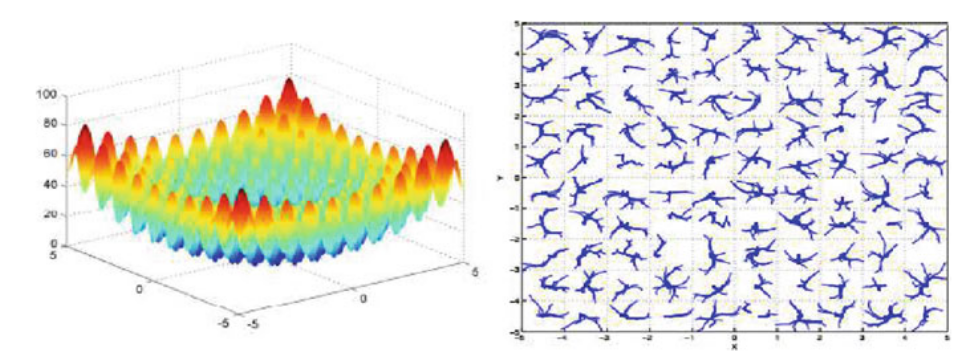}
    \caption{Rastrigin function Convergence and Emergence plot}
    \label{fig:fig5}
\end{figure}

The BMO algorithm was formulated to capture the local optima of multimodal functions. To check the efficacy of the algorithm, it was applied on the 3-D benchmark multimodal functions. Initially, the bflies are randomly deployed on the entire search space of 3-peaks standard benchmark function. Fig~\ref{fig:fig3}. a, b, c shows the 3-peaks function, the emergence of all Bflies and the UV convergence at the end of all iterations respectively. In the same line of thought, the BMO has applied with two other benchmarks functions viz. Schwefel (peaks - 15) (Fig~\ref{fig:fig4}) and Rastrigin (peaks - 100) (Fig~\ref{fig:fig5}) to capture all peaks.

\subsection{Bflybots: Multi-robotic Platform for Implementing Butterfly Mating Phenomenon}

To conduct the real-time experiments on the Butterfly Mating Optimization (BMO) algorithm, a mobile robot namely Bflybot was designed \cite{b20}. The Bflybot will be able to sense/read the light intensity values, distribute the information to all other Bflybots, choose the local mate (l-mate) and finally move towards the chosen l-mate in the pre-defined experimental work space. For all the experiments, the light sources are considered as the signal sources for detection in 2 dimensional work space. The Arduino Uno (ATmega328P) microcontroller is programmed to interface all the sensors for processing the inputs received from the other bots and control the movement with predefined step size.

As mentioned in the BMO algorithm, initially the bflybot needs to sense/read the intensity at their position in each iteration. The Light Dependent Resistor (LDR) used to read the light intensity ahead of it. To increase the effective reading of intensity at present location, the LDR sensor rotated 360 degrees using a stepper motor and took the max value of continuous readings. Upon reading the intensity at current position, each Bflybot updates the UV (intensity) value at the ith iteration using Eqn. 1. The PS2 mouse based odometry protocol measures the coordinates at instant locations of each Bflybot. Each Bflybot will share their UV (by following Eqn. 2) and position values to other bflybots through a wireless Zigbee module (S2C) (This module can be configured in AT mode using the XCTU software tool). The L-mate selection process for each Bflybot is done by substituting the UV values received from other Bflybots in the Eqn. 4. Finally, the Bflybot will move towards the chosen L-mate with a predefined step size. The rotation towards the direction of chosen L-mate is controlled by the Accelerometer (MPU6050). Also, the motor shield (L293D) is used to drive the DC motors towards the L-mate. Every Bflybot deployed in the defined work space will exhibit these characteristics to mimic the BMO algorithm to colocate at the signal source (here light source).

\textbf{Single source detection with various initial placements and stepsizes}

To do the experiments with a single light source \cite{b6}, a white background floor with little roughness is chosen and which is considered an arena for the real-time experiments. In the middle of the arena, A Compact Fluorescent Lamp (CFL) of 14 Watts was placed. Such that the luminescence will be equally distributed in all the directions. The workspace (arena) considers two circles. The former one is the outer circle, which is 180cm diameter for better visualization of Bflybots working phenomena and the latter one is the inner circle with 50 cm. Entering into the inner circle indicates the localization or detection of the light score. The  workspace is divided into four equal parts and each bot will place in each individual part for the equal distribution of the Bflybots. And the left most corner is used as an origin (0, 0) for the workspace. Such that Bflbots knows their initial positions in the XY-plane. PS2 mouse is used to update the Bflybot positions after each iteration. Most often convergence of Bflbots rely on the initial positions of the bots placements. If the bots placed near to the source location, they can converge in minimum elapsed time or else if the bots placed far away from the source then the convergence will increase gradually. 

After each iteration, each Bflybot has to move a certain distance called step-size. For initial experiments we have set the step-size as 10cm and various experiments were performed by varying the step-sizes also. By assigning the small step-size leads to travels a longer path with increase in the convergence time. In contrast to that, giving the larger step-size leads to poor accurate convergence by decreasing the search time. After several experiments, we have identified that 10cm is the ideal step-size to get the accurate convergence. 

\begin{figure}
    \centering
    \includegraphics[width=7in]{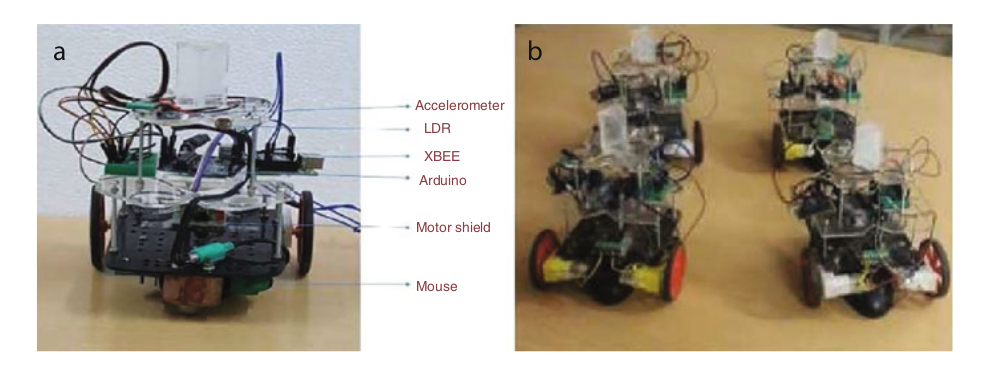}
    \caption{ (a) Bflybot architecture				   (b). Bflybot Swarm}
    \label{fig:fig6}
\end{figure}

The mobile robotic swarm of four Bflybots were prepared, which mimicked the Bflies in the BMO \cite{b6}. The multi-robotic platform is planned to detect locations of multiple light sources. The Bflybots have designed and prepared to perform the four phases of the BMO algorithm mentioned in Section-\ref{sec:sec2}. In brief, UV (light source) updating, distribution and L-mate selection done using the Light Dependent Resistor (LDR) sensor and ZigBee module, movement was done using an accelerometer and mouse-based X-Y position calibration. Fig~\ref{fig:fig6}. a, b shows the Bflybot with components specified and Bflybot swarm respectively.

\subsection{Bflybots for detection and co-locating the static light sources}

\begin{figure}
    \centering
    \includegraphics[width=7in]{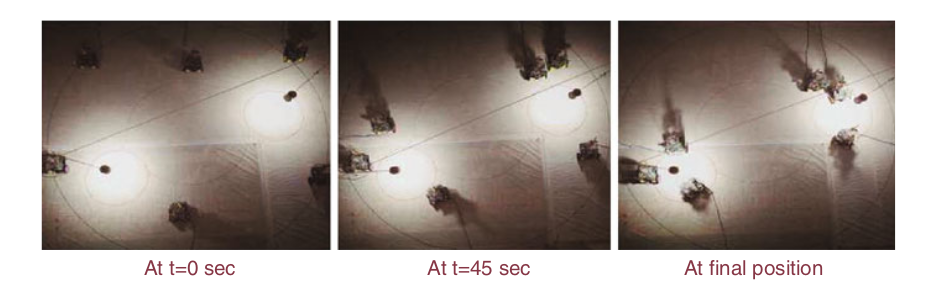}
    \caption{Similar color sources convergence at various intervals of time}
    \label{fig:fig7}
\end{figure}

For conducting the experiments on multiple sources \cite{b7}, the work space was prepared with an outer circle 115cm radius and two inner circles with 25cm radius as shown in Fig~\ref{fig:fig7}. Two light sources (with 14 watts) are hung in such a way that the illuminated light of each light source, situated on inner circles (indication that the inner circles are the peaks). The workspace was divided into four equal quadrants and in each quadrant an equal number of bots were placed randomly with reference to origin. This distribution and deployment of bots in random places at each quadrant guarantee that maximum coverage of work space. As shown in Fig~\ref{fig:fig7}, the Bflybots follow the steps involved in the BMO and converge at each light source (at inner circles) placed in the work space. Here, the initial placement of the Bflybots will impact the number of bots converging at each light source. Even Though, the localization is limited to local regions (converging to nearest light source), the communication between the bflybots is global (every bflybot shares their information with all other bflybots). This observation is confirmed from the experiments that the Bflybot near to the local source also moved towards another source at some particular point of time. They were not confined to a particular portion but instead they act globally in the workspace based on the intensity distribution.

Similarly, the experiments were extended to detect the multiple light sources with differences in intensity levels \cite{b7}. To achieve that variation in intensity levels, two light sources with Red and Blue color lamps (with 14 watts) are chosen. As mentioned above, the bflybots are deployed in the work space randomly. Upon certain iterations the bots converged at the two source locations. From these experiments, it is evident that the Bflybots are able to colocate at the different peaks (able to identify all the light sources with various intensity levels). With this inspiration, we made an attempt to detect the dynamically varying signal sources. For this, the firstmost step is to simulate the proposed BMO algorithm with the benchmarks functions by varying the peaks with respect to time. For simplicity, one of the standard benchmark three peaks functions is taken into consideration. The following subsection describes how the varying peaks can be captured at certain time instances.

\subsection{Bflybots for chasing and interception of dynamic light sources}

To cope up with the real-time applications which are dynamic in nature, BMO algorithm not only detects the static sources but also captures the dynamically varying sources and is able to adapt the changes in the environment at every instant of time. BMO algorithm is applied to a dynamically varying 3-peaks function to simulate the results while capturing the peaks varying horizontally as shown in Fig~\ref{fig:fig8}. The peaks can vary horizontally in each iteration using the Eqn. 5

\begin{figure}
    \centering
    \includegraphics[width=7in]{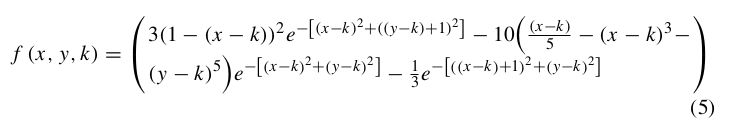}
\end{figure}

Where, k is the constraint to shift the peaks horizontally.

\begin{figure}
    \centering
    \includegraphics[width=7in]{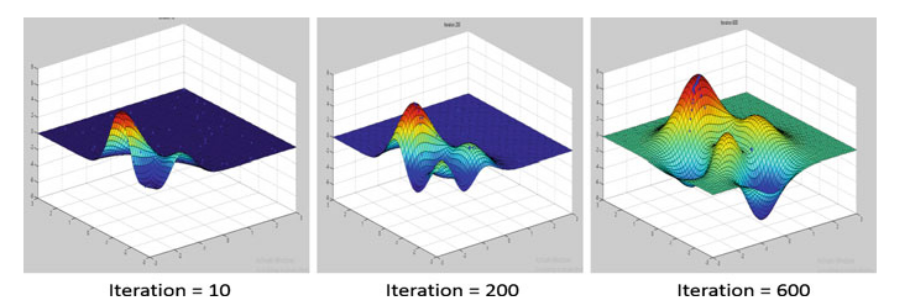}
    \caption{ 3-peaks horizontally varying case}
    \label{fig:fig8}
\end{figure}

\begin{figure}
    \centering
    \includegraphics[width=7in]{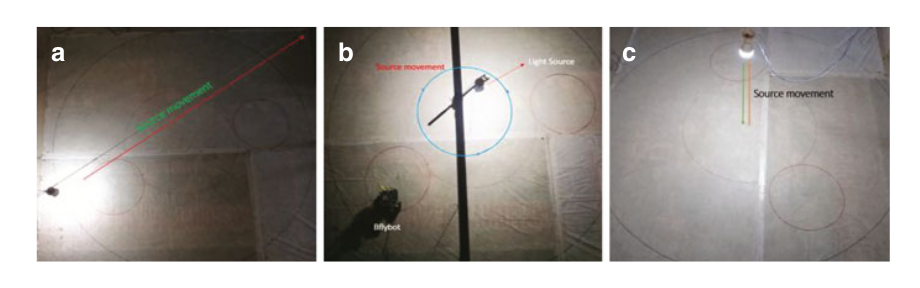}
    \caption{(a). Linear movements	(b). Circular movement 		       (c). Up-down movement}
    \label{fig:fig9}
\end{figure}

These simulation works were extended to implementation with swarm robots namely Bflybots. The source(s) moved in different movements like linear, circular, up and down etc., to create the dynamic environment in the work-space as shown in Fig~\ref{fig:fig9}.

\begin{figure}
    \centering
    \includegraphics[width=7in]{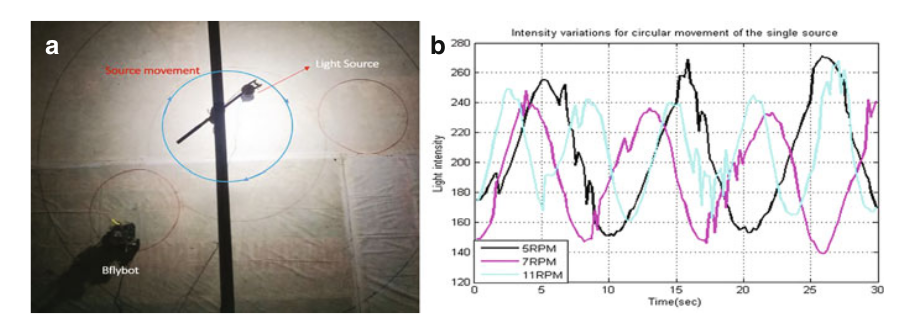}
    \caption{(a). Circular movement of light source 	(b). Light intensity variations}
    \label{fig:fig10}
\end{figure}

Initially experiments on dynamically varying source(s) started by analysing the light sensing capability of Bflybot. Here, a Bflybot was placed in the work-space and rotated the light-source in a circular path as shown in Fig~\ref{fig:fig10}. a. The Bflybot continuously captures the light intensity at the present position. Fig~\ref{fig:fig10}. b, shows the intensity variations on various Revolutions Per Minute (RPM) of the source movement. One crucial observation is that the intensity value depends on the RPM of source movement. If the light source moves with more RPM then the intensity at Bflybot may decrease (due to light sensor inaccuracy in capturing intensity variations).

\begin{figure}
    \centering
    \includegraphics[width=7in]{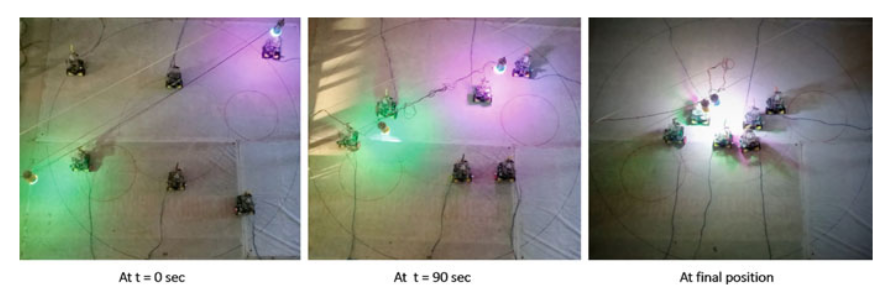}
    \caption{ Linearly varying multiple signal sources with different color}
    \label{fig:fig11}
\end{figure}

\begin{figure}
    \centering
    \includegraphics[width=7in]{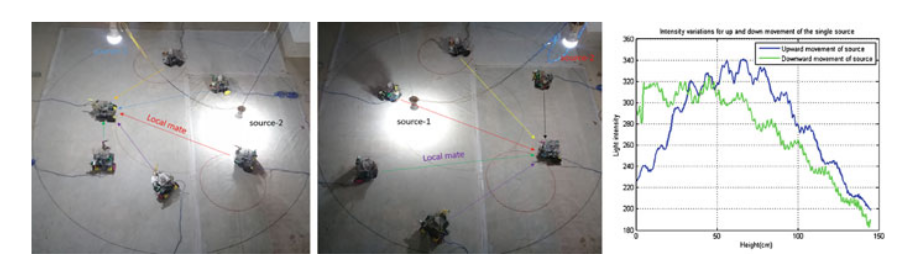}
    \caption{Responsive behavior of Bflybots whilst moving light sources up and down}
    \label{fig:fig12}
\end{figure}

To extend the same experiments on dynamically varying multi-source localization \cite{b8}, two different color sources (pink and green) were chosen as shown in Fig~\ref{fig:fig11}. Here, the Bflybots capture the light intensity variations of two sources (moving towards each other) and chase till localization at dynamic source locations. One of the crucial observations here is that the Bflybots’ are more likely to choose their L-mates which are nearer to the light sources at the particular time instance. Also, the Bflybots follow the light source movement like chasing instead of localizing at the light source location. This indicates that the BMO algorithm is likely to adapt to the new changes in the work space. In the same line of thought, experimenters were performed by moving the multiple sources in up and down movement as shown in Fig~\ref{fig:fig12}. Here, the Bflybots exhibit the through and fro motion to up and down movement of light sources. As shown in Fig~\ref{fig:fig12} a \& b, the directed arrows signifies the movement of Bflybots towards their chosen L-mate.

Using the multi-robotic platform, the above mentioned experiments were conducted to verify the rendezvous phenomena to detect and co-locate the static and dynamic light sources \cite{b7}\cite{b8} and come up with the following conclusions.

\begin{enumerate}
    \item  When two light sources have similar (Fig~\ref{fig:fig7}) or different intensity levels, the number of Bflybots convergence at sources depends on their initial positions and step-size of the individual Bflybots \cite{b6}. 
 \item As the localization of Bflybots was limited to the local region but the communication among Bflybots was global \cite{b7}.
\item Authors of bhattiprolu et. al \cite{b8} confirmed that BMO was able to detect dynamically varying sources (Fig~\ref{fig:fig9}) and experiments were performed to detect manoeuvring (circular and up-down) and non-manoeuvring (linear) movements of the light sources.
\end{enumerate}

\subsection{Comparisons with Existing Optimization Algorithms}

To the best of our knowledge, by using the above-mentioned conclusions, we have identified the following comparisons with existing optimization algorithms.

\begin{enumerate}
    \item Particle Swarm Optimization (PSO), Glowworm Swarm Optimization (GSO), Ant Colony Optimization (ACO) are the popular optimization algorithms, but they have many challenges to implement in dynamic environments. Unlike the above algorithms, the proposed BMO algorithm is able to implement most of the challenges that are involved in the dynamic environment also.
\item Compared to PSO and GSO, the BMO algorithm has less tuning parameters, which reduces the time consumption for the convergence.

\end{enumerate}

\section{PROSPECTIVE AND WORKS-IN-PROGRESS}
\label{sec:sec4}

We have identified several real-time applications to check the BMO efficacy on them

\subsection{Mid-Sea ship detection using satellite image data set}

Ship detection from satellite images can be considered as a simple detection of high intensity (in grayscale) points against a noisy background. This problem has a very much impact on identification of illegal oil spills and monitoring maritime traffic in the fisheries and commercial transportation civil sectors. Also, these sectors have the largest industries across the world and a typical control system is needed in such a way that it can monitor continuously to identify any threats in advance. However, in reality the noise (like clouds, wave crests etc) present in the image can be detected as a ship based on the colour, shape and brightness. The rate of false positive rate will increase with respect to the noise present in the image. Also, there can be a possibility that the ships can be joined together (the shape will fail in this case as the shape is the aggregation of all ships). To tackle these challenges in real-time scenarios, several techniques were proposed. In those, the most state of art models are detailed below.

Basically two types of remote sensing techniques proposed for ship detection: the former one is Synthetic Aperture Radar (SAR) with capacity to image day and night under most meteorological conditions, became the state of the art technique for ship detection \cite{b18}. In the subsequent approach, authors of Inggs and Robinson \cite{b19} investigated the use of radar range profiles as target signatures for the identification of ship targets and neural networks for the classification of these signatures. Majority of the methods used the SAR data to detect the ship targets in various scenarios.

To complement existing regulations, Corbane et.al \cite{b14}, implemented an operational ship detection algorithm by using the high spatial resolution optical imagery, especially the fishing control system. Along with, comprehensive study performed on the recent progress on the detection model and existing prototypes to classify the small targets.  Also testing performed on panchromatic SPOT5 imagery by considering into account the environmental and fishing context in French Guiana.

With the help of optical flow and saliency methods, Deng et.al \cite{b15} performed several experiments and designed an algorithm to detect ship from optical satellite images, which detect more than one ship targets in the complex dynamic sea background and shows effective results in reducing the false positive rate by compared to the other traditional methods. This research group also discussed detection of moving targets in the image, those are highlighted with the help of classical optical flow method. By using the state-of-art saliency method, dynamic waves are restrained. 

To solve the problems arising with the narrow width of the ship, Yang et.al \cite{b16} modified the Dense Feature Pyramid Network (DFPN). To predict the minimum circumscribed rectangle of the object, they have designed a rotation anchor strategy. Such way the redundant detection region will be reduced and correspondingly the recall will be improved. To maintain the completeness of semantic and spatial information, the research group also proposed a multi-scal ROI. Experiments on remote sensing images from Google Earth for detection show that our R-DFPN based methods outperforms all the existing methods' performance.

For detecting the small vessels on very high resolution satellite images by applying the modern computer vision techniques, Mattyus et.al \cite{b17} proposed a novel fast method. The detector is able to concentrate on vessels shorter than 20m. The vessels are mostly not equipped with AIS and they don’t have special interest for authorities.  Those can hardly be detected by SAR vessel detection methods. This detection method is effective above 8 meter of length. Below this length a wave can have a similar shape as a small vessel. The detector runs fast approximately in a minute for a 16873 × 14684 pixels image on a modern multi-core computer, thus enabling near real time application, i.e, one hour from image acquisition to end user. 

\begin{figure}
    \centering
    \includegraphics[width=7in]{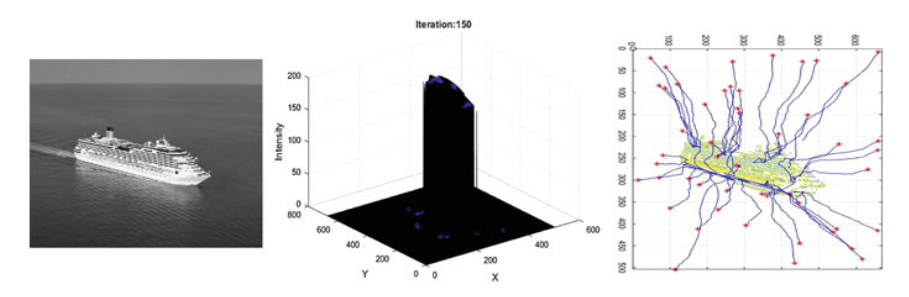}
    \caption{Ship Detection Simulations }
    \label{fig:fig13}
\end{figure}

To extend the scope of BMO for real-time applications, the algorithm applied to ship images at mid-sea from Kaggle dataset \cite{b9} and simulated the co-locating nature of BMO. Initially, BMO is applied to normal images by converting them into gray-scale and observed distraction of co-locations in search space due to the noise present in the images. For simplicity, the images which have less noise are considered for the simulations with the BMO algorithm. Even Though the images are converted to the grayscale, the region where the ship is actually located has some variations in values. To avoid such variations, the normalization technique is applied to sharpen the peak of the ship and ground the other noise in search space as shown in Fig. 14b. Initially the bflies randomly deployed in the normalized surface by considering the peak as the ship. Upon certain iterations, we obtain the co-location of Bflies at the edges of the ship. Further the boundary can be drawn using the adjacent points located at the edge of the ship. Fig.\ref{fig:fig13} shows the simulations results and co-location of Bflies at edges of the ship. Later, similar experiments are conducted on multiple ships present in the image to obtain the collocation of bflies at the edge of each ship. Compared to the state of art models for ship detection, the BMO algorithm identifies the clouds as ships because of the noise. But the algorithm is able to identify the ships present in the images accurately.

\subsection{ Butterflies for water and aerial search applications}

\begin{figure}
    \centering
    \includegraphics[width=7in]{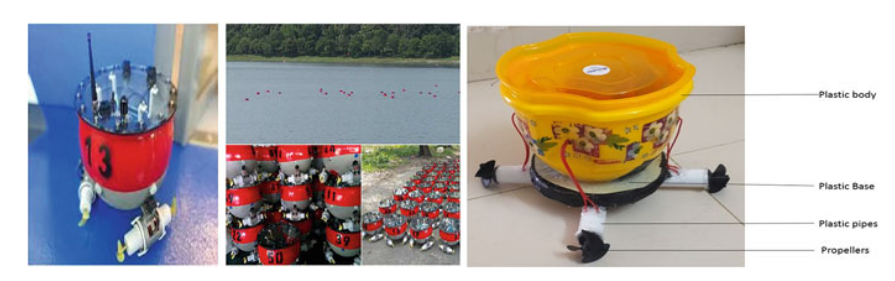}
    \caption{Swarm of Buoys developed in Bouffanais’s Lab and Waterbot}
    \label{fig:fig14}
\end{figure}

Most of the real-time applications exist in 3-dimensional work space. To simulate and implement the BMO algorithm in a 3-dimensional search space, we move a step forward to solve two kinds of applications. Among those one similar to the Buoys developed in Bouffanais’s lab shown in Fig~\ref{fig:fig14}a. In this the bots are designed to move on top of water for collection of garbage floating on water. With the same length of two plastic pipes are fixed to the base of the robot (where the two plastic pipes are orthogonal to each other). A lightweight plastic box to build the body of the robot and fixed all the components of the bot inside the box. Each end of these two pipes were fixed with the propellers and connected to the motors with 25000 rpm as shown in Fig~\ref{fig:fig14}b. The microcontroller Arduino Uno (ATmega328P) and motor drivers were placed in the body to control the movement of the waterbot. As described in the last phase of the BMO algorithm, the movement towards the L-mate of water can be controlled by the motor drivers. Also, the rotation taken by the waterbot while moving towards the L-mate, can be controlled by the Accelerometer (MPU6050). For communication with other bots, the XBEE module is used.

\begin{figure}
    \centering
    \includegraphics[width=7in]{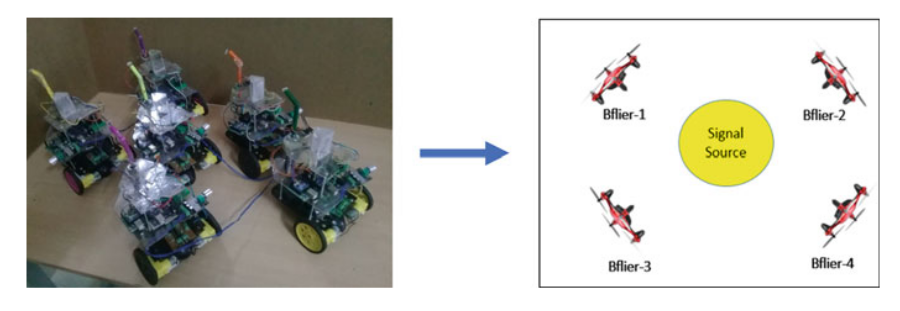}
    \caption{Transformation from multirobotic platform to quad-rotors platform}
    \label{fig:fig15}
\end{figure}
Also, we started simulating the experiments in Player-Stage software and then intended to use a real Quad-rotor swarm in a 3-D environment. Fig~\ref{fig:fig15} gives the intuition behind the transformation from multi-Bflybot swarm with multi Quad-rotor platform. As shown in Fig~\ref{fig:fig15}, we are planning to detect a signal source with the help of four Quad rotors, where each one is similar in design and working phenomenon characteristics. In a way which leads to follow the swarm behavior. Experiments with multi- quad rotors especially useful in applications such as detecting the fire locations in the forest area. Further, the similar can be performed to detect the multiple sources and dynamically varying signal sources detection purpose too. In case of fire location detection also, the flame can be spread across the forest in such cases the locations of global maximum varies with respect to time. In such cases, BMO effectively works on detecting the dynamically varying signal source locations. 

\begin{figure}
    \centering
    \includegraphics[width=7in]{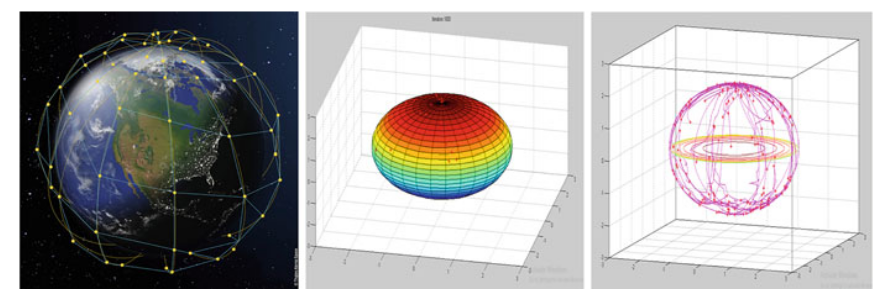}
    \caption{Satellite constellation for BMO application}
    \label{fig:fig16}
\end{figure}

\subsection{Simultaneous Earthquake prediction and detection with satellite constellation}

In this work, it is planned to take seismograph measures and other image-based cues from satellite constellations periodically and will apply the BMO paradigm to predict the possibility of upcoming earthquakes at specific locations on earth. Initially, the benchmark function (i.e Sphere) was drawn from the Eqn. 6 and randomly deployed the bflies on the surface. Also, the satellite data incorporated on the sphere surface and considered as the fitness. Here, the  north and south poles are considered to be peak and valley respectively. At certain iterations, it is found that the Bflies are co-locating to the virtual north pole on the sphere, which is nothing but a peak on the sphere. From these simulations, it is evident that the BMO algorithm is able to identify the change in fitness on the sphere surface. With this intuition, similar experiments will be conducted on the seismograph measures and other image-based cues to predict the earthquake on earth. Fig~\ref{fig:fig16} shows the simulated results of co-locations on earth using satellite data.

\begin{equation}
                     x^{2}+y^{2}+z^{2} \le r^{2}          
\end{equation}

Where r is the radius of the sphere within the domain range [0, 2].


\section{Analysis and Discussion}
\label{sec:sec5}

This chapter presents the various simulation results and the corresponding implementations in the multirobotic platform for the BMO algorithm. The experimental
results indicate the efficacy of the BMO in detecting static and dynamically moving light source(s) \cite{b6, b7, b8}. Several experiments concluded on varying step-size and initial placements \cite{b6} of Bflybots and observed that these variations played a crucial role in light source(s) detection. Varying initial placements and step-sizes affect the convergence in terms of time
and smoothness \cite{b6}, whereas varying step-size leads to the variation in the execution time and speed of the Bflybots \cite{b6}. We have
identified suitable comparisons (as mentioned in the section 3.6) with other popular optimization algorithms. Unlike other popular algorithms, BMO can be applied to
dynamic environments also. In BMO, the number of tuning parameters was less compared to other algorithms. And we have also observed that the convergence of BMO is local but the communication between the Bflybots is done globally.

We also planned to take a step, to present the ongoing and prospective works. In those, we are planning to embed the model into a multi-quad rotor swarm, which could be the ultimate reality and suits many practical applications \cite{b11}. The trails on dynamic sources could lead to tsunami location identification too. BMO can be applicable to both water bot and Earthquake prediction scenarios and some kind of object detection applications by considering the object as a peak.

\section{RELATED WEBLINKS}

\begin{enumerate}
    \item \url{https://www.enchantedlearning.com/subjects/butterfly/allabout/index.shtml}
\item \url{http://www.butterflyzone.org/}
\item \url{https://www.monarchwatch.org/biology/pred1.htm}

\end{enumerate}

\end{document}